\documentclass[10pt, a4paper]{article}

\usepackage{lrec-coling2024} 

\usepackage{times}
\usepackage{latexsym}

\usepackage[T1]{fontenc}

\usepackage[utf8]{inputenc}

\usepackage{microtype}

\usepackage{inconsolata}

\usepackage{graphicx}
\usepackage{amsmath}
\usepackage{amsfonts}
\usepackage{amssymb}
\usepackage{array}
\usepackage{amsthm}
\usepackage[ruled,vlined]{algorithm2e}
\usepackage{booktabs}
\usepackage{multirow}
\usepackage{multicol}

\usepackage{xurl}

\usepackage{enumitem}

\theoremstyle{definition}
\newtheorem{definition}{Definition}[section]

\usepackage[noend]{algpseudocode}

\usepackage{subfiles} 
\usepackage{courier}

\renewcommand{\d}{\mathrm{d}}

\newcolumntype{P}[1]{>{\centering\arraybackslash}p{#1}}

\title{A Differentiable Integer Linear Programming Solver for Explanation-Based Natural Language Inference}

\name{Mokanarangan Thayaparan$^{1,3}$, Marco Valentino$^{3}$, Andr\'e Freitas$^{1,2,3}$\\
$^{1}$Department of Computer Science, University of Manchester, UK\\  
$^{2}$ National Biomarker Centre, CRUK-MI, University of Manchester, UK\\
$^{3}$Idiap Research Institute, Switzerland \\ }

\address{mokanarangan.thayaparan@manchester.uk, marco.valentino@idiap.ch, andre.freitas@idiap.ch\\}

\abstract{
Integer Linear Programming (ILP) has been proposed as a formalism for encoding precise structural and semantic constraints for Natural Language Inference (NLI). 
However, traditional ILP frameworks are non-differentiable, posing critical challenges for the integration of continuous language representations based on deep learning. In this paper, we introduce a novel approach, named Diff-Comb Explainer, a neuro-symbolic architecture for explanation-based NLI based on Differentiable BlackBox Combinatorial Solvers (DBCS). 
Differently from existing neuro-symbolic solvers, Diff-Comb Explainer does not necessitate a continuous relaxation of the semantic constraints, enabling a direct, more precise, and efficient incorporation of neural representations into the ILP formulation. Our experiments demonstrate that Diff-Comb Explainer achieves superior performance when compared to conventional ILP solvers, neuro-symbolic black-box solvers, and Transformer-based encoders. Moreover, a deeper analysis reveals that Diff-Comb Explainer can significantly improve the precision, consistency, and faithfulness of the constructed explanations, opening new opportunities for research on neuro-symbolic architectures for explainable and transparent NLI in complex domains. 
 \\ \newline  }

\begin{document}

\maketitleabstract

\section{Introduction}
\label{sec:intro}

\begin{figure}[t]
    \centering
    \includegraphics[width=0.5\textwidth]{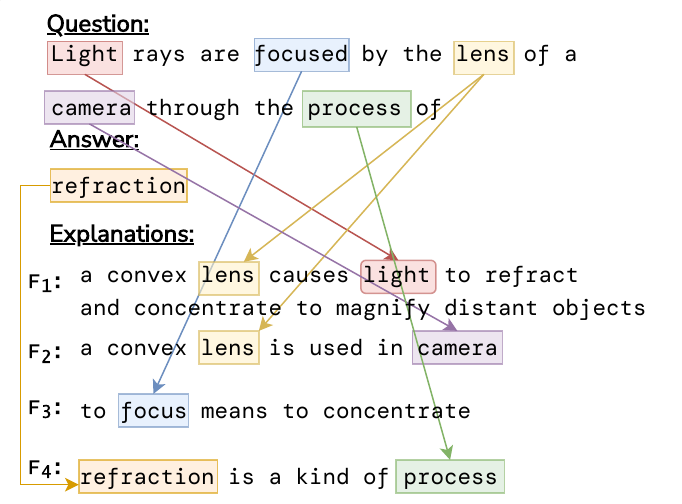}
    \caption{Example of a hypothesis (i.e., question + answer) and an explanation graph constructed via an ILP-based NLI model ~\citep{xie2020worldtree,jansen2018worldtree}.}
    \label{fig:comb_end_to_end_example}
\end{figure}

Given a question and a possible candidate answer expressed in natural language (which we refer to as \emph{hypothesis}), ILP-based Natural Language Inference (NLI) aims to construct an explanation graph consisting of interconnected sentences to support and deduce the hypothesis (Figure~\ref{fig:comb_end_to_end_example}). 

To contrast the lack of interpretability of large deep learning models, ILP has been proposed as a formalism to build transparent and explainable NLI models, able to leverage explicit semantic constraints to address downstream inference tasks. ILP-based solvers, in fact, provide a viable mechanism to encode explicit and controllable assumptions about the structure of the explanations \citep{khashabi2018question,khot2017answering,khashabi2016question,thayaparan2020survey}. 

However, existing ILP frameworks are \textit{non-differentiable} and cannot be integrated as part of a broader deep learning architecture~\citep{DBLP:journals/corr/abs-2105-02343,poganvcic2019differentiation}. Therefore, these approaches have been so far limited by the exclusive adoption of hard-coded heuristics and cannot be easily optimised end-to-end to achieve performance comparable to deep learning counterparts~\citep{thayaparan2021explainable,khashabi2018question}. At the same time, existing solutions for the integration of combinatorial solvers and deep learning require a continuous approximation of the semantic constraints, which inevitably lowers the expressive power of the ILP formulation and, as a consequence, the precision in the downstream inference process \cite{diff-explainer}.

To address these challenges, this paper proposes an end-to-end methodology for the integration of \textit{Differentiable BlackBox Combinatorial Solvers} (DBCS)~\citep{poganvcic2019differentiation} into explanation-based NLI architectures.  DBCS, in fact, enables an efficient approximation of the gradient for end-to-end optimisation, without the need for a continuous relaxation of the original ILP formulation. Specifically, we demonstrate how a combinatorial optimization solver can be adopted as a composable building block in a broader NLI architecture with Transformers \cite{devlin2019bert}.


In summary, the contributions of this paper are as follows:



\begin{enumerate}[noitemsep,leftmargin=*]
    \item We propose \textit{Diff-}Comb Explainer, a novel ILP-based NLI solver that combines a differentiable black box combinatorial solver with Transformer-based encoders, enabling end-to-end differentiability without the need for a continuous relaxation of the ILP formulation.
    \item We empirically demonstrate that {Diff-}Comb Explainer achieves superior performance on both explanation generation and answer selection when compared to differentiable and non-differentiable counterparts.
    \item We demonstrate that thanks to the preservation of the original ILP formulation, \textit{Diff-}Comb Explainer better reflects the underlying explanatory inference process leading to the final answer prediction, outperforming existing combinatorial solvers in terms of faithfulness and consistency.\footnote{Code and data available at \url{https://github.com/neuro-symbolic-ai/diff_comb_explainer}}
\end{enumerate}

\section{Related Work}

\paragraph{Constraint-based multi-hop inference} ILP has been applied for structured representation~\citep{khashabi2016question} and over semi-structured representation extracted from text~\citep{khot2017answering,khashabi2018question}. Early approaches were unsupervised. However, recently \citet{thayaparan2021explainable} proposed the ExplanationLP model optimised towards answer selection via Bayesian optimisation. ExplanationLP was limited to fine-tuning only nine parameters and used pre-trained neural embedding. \textit{Diff-}Explainer~\citep{diff-explainer} was the first approach to integrate constraints into a deep-learning network via Differentiable Convex Optimisation Layer~\citep{agrawal2019differentiable} by approximating ILP constraints using Semi-definite programming.~\citep{lovasz}. Differently from our approach, however, \emph{Diff}-Explainer requires a continuous approximation of the semantic constraints, a feature that inevitably affects the precision in the downstream inference process.

\paragraph{Hybrid reasoning with Transformers} ~\citet{clark2021transformers} proposed ``soft theorem provers” operating over explicit theories in language. This hybrid reasoning solver integrates natural language rules with transformers to perform deductive reasoning.  \citet{saha2020prover} improved on top of it, enabling the answering of binary questions along with the proofs supporting the prediction. The multiProver~\citep{saha2021multiprover} evolves on top of these conceptions to produce an approach that is capable of producing multiple proofs supporting the answer. While these hybrid reasoning approaches produce explainable and controllable inference, they assume the existence of natural language rules and have only been applied to synthetic datasets. On the other hand, our approach does not require extensive rules set and can tackle complex scientific and commonsense NLI problems.

\section{\textit{Diff-}Comb Explainer: Differentiable Integer Linear Programming Solver for Explanation-Based Natural Language Inference}
\label{sec:comb_approach}

\begin{figure*}[t]
    \centering
    \includegraphics[width=\textwidth]{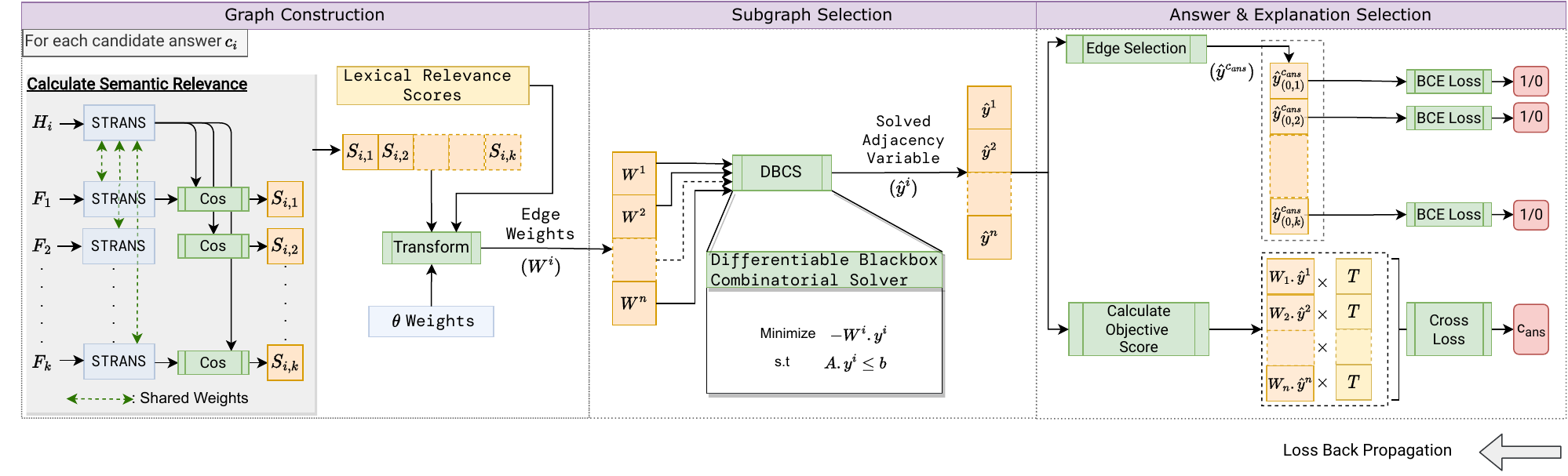}
    \caption{End-to-end architectural diagram of \textit{\textit{Diff-}Comb Explainer}. The integration of Differentiable Blackbox Combinatorial solvers will result in better explanation generation and answer prediction.}
    \label{fig:end_to_end_comb}
\end{figure*}

ILP-based NLI is typically applied to multiple-choice question answering problems~\citep{khashabi2018question,khot2017answering,khashabi2016question,thayaparan2021explainable}. Given a Question ($Q$) and a set of candidate answers $C=\{c_1,~c_2,~c_3,~\dots,~c_n\}$ the aim of the task is to select the correct answer $c_{ans}$. 

In order to achieve this, ILP-based approaches convert question-answer pairs into a list of hypotheses $H=\{h_1,~h_2,~h_3,~\dots,~h_n\}$ (where $h_i$ is  the concatenation of $Q$ with $c_i$) and typically adopt a  retrieval model (e.g: BM25, FAISS~\citep{DBLP:journals/corr/JohnsonDJ17}),  to select a list of candidate explanatory facts $F = \{f_{1},~f_{2},~f_{3},~\dots,~f_{k}\}$. Then construct a weighted graph $G = (V,E,W)$ with edge weights $W: E~\rightarrow~\mathbb{R}$ where $V=~\{\{h_i\}~\cup~F\}$, edge weight $W_{ik}$ of each edge $E_{ik}$ denote how relevant a fact $f_k$ is with respect to the hypothesis $h_i$.  

Given this premise, ILP-based NLI can be defined as follows~\citep{diff-explainer}:

\begin{definition}[\textit{ILP-Based NLI}]
Find a subset $V^* \subseteq V$, $h \in V^*$ and $E^* \subseteq E$  such that the induced subgraph $G^* = (V^*,E^*)$ is connected, weight $W[G^* = (V^*,~E^*)] := \sum_{e \in E^*} W(e)$ is maximal and adheres to set of constraints $M_c$ designed to emulate multi-hop inference. The hypothesis $h_i$ with the highest subgraph weight $W[G^* = (V^*,~E^*)]$ is selected to be the correct answer $c_{ans}$.
\end{definition}

As illustrated in Figure~\ref{fig:end_to_end_comb}, \textit{Diff-}Comb Explainer has 3 major parts: \textit{Graph Construction}, \textit{Subgraph Selection} and \textit{Answer/Explanation Selection}. In \textit{Graph Construction}, for each candidate answer $c_i$ we construct a graph $G^i = (V^i,~E^i,~W^i)$ where $V^i =  \{h_i\} \cup \{ F \}$ and the weights $W^i_{ik}$ of each edge $E^i_{ik}$ denote how relevant a fact $f_k$ is with respect to the hypothesis $h_i$. These edge weights ($W^i_{ik}$) are calculated using a weighted ($\theta$) sum of scores calculated using transformer-based ($STrans$) embeddings and lexical models.

In the \textit{Subgraph Selection} step, for each $G^i$, Differentiable Blackbox Combinatorial Solver (DBCS) with constraints are applied to extract subgraph $G^*$. In this paper, we adopt the constraints proposed for ExplanationLP~\citep{thayaparan2021explainable} for ensuring a fair comparison between different models. ExplanationLP explicit abstraction by grouping facts into abstract ($F_{A}$) and grounding ($F_{G}$). Abstract facts are the core scientific statements that a question is attempting to test, and grounding facts link concepts in the abstract facts to specific terms in the hypothesis (e.g. taxonomic relations). For example, in Figure~\ref{fig:comb_end_to_end_example} the core scientific fact is about the nature of convex lens and how they refract light ($F_1$). Grounding Facts $F_2, F_3, F_4$ are usually is-a relations that help to connect abstract concepts to concrete instances in the hypothesis.

Finally, in \textit{Answer/Explanation Selection} the model predicts the final answer $c_{ans}$ and relevant explanations $F_{exp}$. During training time, the loss is calculated based on gold answer/explanations to fine-tune the transformers ($STrans$) and weights ($\theta$). The rest of the section explains each of the components in detail.

\subsection{Graph Construction}

In order to facilitate the construction of inference chains, the retrieved facts $F$ are classified into \textit{grounding} facts $F_{G} = \{f^g_{1}, f^g_{2}, f^g_{3}, ..., f^g_{l}\}$ and abstract facts $F_{A} = \{f^a_{1}, f^a_{2}, f^a_{3}, ..., f^a_{m}\}$  such that $F=F_{A} \cup F_{G}$ and $l + m = k$ \cite{thayaparan2021explainable}.
Here, We use two relevance scoring functions to calculate the edge weights: semantic and lexical scores. We use a Sentence-Transformer (STrans)~\citep{reimers2019sentence} bi-encoder architecture to calculate the semantic relevance. The semantic relevance score from STrans is complemented with the lexical relevance score. The semantic and lexical relevance scores are calculated as follows:

\paragraph{Semantic Relevance ($s$)}: Given a hypothesis $h_i$ and fact $f_j$ we compute sentence vectors of $\vec{h_i}^{\,} = STrans(h_i)$ and $\vec{f_j}^{\,} = STrans(f_j)$ and calculate the semantic relevance score using cosine-similarity as follows:
    \begin{equation}
        s_{ij} = S(\vec{h_i}^{\,},~\vec{f_j}^{\,}) =  \frac{\vec{h_i}^{\,} \cdot \vec{f_j}^{\,}}{ \|\vec{h_i}^{\,} \|\|\vec{f_j}^{\,}\|}
    \end{equation}
    
\paragraph{Lexical Relevance ($l$)}: The lexical relevance score of hypothesis $h_i$ and $f_j$ is given by the percentage of overlaps between unique terms (here, the function $trm$ extracts the lemmatized set of unique terms from the given text):
    \begin{equation}
        l_{ij} = L(h_i,~f_j) = \frac{\vert trm(h_i) \cap trm(f_j) \vert}{max(\vert trm(h_i)\vert, \vert trm(f_j) \vert)}
    \end{equation}
    
Given the above scoring function, we construct the edge weights matrix ($W^i$) for each hypothesis $h_i$ as follows:

\begin{equation}
\small
W^i§§_{jk} = 
\begin{cases}
-\theta_{gg}l_{jk} & (j,k)\in F_{G}\\
-\theta_{aa} l_{jk}  & (j,k) \in F_{A}\\
\theta_{ga}l_{jk}  & j \in F_{G}, k \in F_{A}\\
\theta_{qgl}l_{jk} + \theta_{qgs} s_{jk}  & j \in F_{G}, k = {h_i}\\
\theta_{qal}l_{jk} + \theta_{qal}s_{jk} & j \in F_{A}, k = {h_i} \\
\end{cases}
\end{equation}

Here relevance scores are weighted by  $\theta$ parameters which are clamped to $[0,1]$.

\subsection{Subgraph Selection}


Given the above premises, the objective function is defined as:

\begin{equation}
	 \min~\qquad -1(W^i \cdot y^i)
\end{equation}

We adopt the edge variable $y^i \in \{0,1\}$ where $y^i_{j,k}$ ($j \neq k$) takes the value of 1 if edge $E^i_{jk}$ belongs to the subgraph and $y^i_{jj}$ takes the value of 1 iff $V^i_j$ belongs to the subgraph. 

Given the above variable, the constraints are defined as follows:

\paragraph{Answer selection constraint} The candidate hypothesis should be part of the induced subgraph:
\begin{equation}
    \begin{aligned}
    \sum_{j~\in~\{h_i\}} y^i_{jj} & = 1 \\ 
    \end{aligned}
\end{equation}

\paragraph{Edge and Node selection constraint} If node $V^i_j$ and $V^i_k$ are selected then edges $E^i_{jk}$ and $E^i_{kj}$ will be selected. If node $V^i_j$ is selected, then edge $E_{jj}$ will also be selected:
\begin{align}
     y^i_{jk} \leq&~y^i_{jj}&& \forall~(j,k) \in E  \\
    y^i_{jk} \leq&~y_{kk}&& \forall~(j,k) \in E \\
    y^i_{jk} \geq&~y_{jj}+y_{kk}-1 && \forall~(j,k) \in E
\end{align}

\paragraph{Abstract fact selection constraint} Limit the number of abstract facts selected to $M$: 

\begin{align}
     \sum_{i} y^i_{jj} \leq&~M &&\forall j \in F_{A}
\end{align}

\subsection{Adopting Differentiable Blackbox Combinatorial Optimisation Solvers}  

The subgraph selection problem can be written in the following ILP canonical form:
\begin{equation} \label{E:BILP}
    \min_{y^i\in \{0,1\}}\, -(W^i \cdot y^i)
		\qquad \text{subject to} \qquad
	Ay^i \le b,
\end{equation}

where $A=[a_1,\ldots,a_m]\in\mathbb{R}^{m\times n}$ is the matrix of constraint coefficients and $b\in\mathbb{R}^m$ is the bias term. The output of the solver $g(W^i)$ returns the \texttt{argmin}$_{y^i\in \{0,1\}}$ of the integer problem.

Differentiable Combinatorial Optimisation Solver~\citep{poganvcic2019differentiation} (DBCS) assumes that $A$, $b$ are constant and the task is to find the $\d L/\d y^i$ given global loss function $L$ with respect to solver output $y^i$ at a given point $\hat y^i=g(\hat W^i)$. However, a small change in $W$ is \textit{typically} not going to change the optimal ILP solution resulting in the true gradient being zero. DBCS provides a theoretical framework that can potentially make any black box combinatorial solver differentiable. In this paper, we adopt the proposed algorithm as following to design the forward/backward pass:

\begin{algorithm}
  \caption{\textit{Forward} and \textit{Backward} for DBCS }\label{alg:two}
  \DontPrintSemicolon
  \SetKwFunction{FMain}{ForwardPass}
  \SetKwProg{Fn}{Function}{:}{}
  \Fn{\FMain{$\hat{W^i}$}}{
        $\hat{y^i}$ = $g(\hat{W})$\;
        \textbf{save} $\hat{W^i}$ and $\hat{y^i}$ for backward pass\;
        \KwRet\ $\hat{y^i}$;
  }
  \;
  \SetKwFunction{FMain}{BackwardPass}
  \SetKwProg{Pn}{Function}{:}{\KwRet}
  \Pn{\FMain{$\frac{\d L}{\d y^i}(\hat y^i)$, $\lambda_{dbcs}$}}{
        \textbf{load} $\hat{W^i}$ and $\hat{y^i}$ for backward pass\;
        $\hat{W^i}$ = $\hat{W^i}$ + $\lambda_{dbcs} \frac{\d L}{\d y^i}(\hat y^i)$\;
        $y^i_{\lambda_{dbcs}}$ = $g(\hat{W^i})$\;
        \KwRet\ $-\frac{1}{\lambda}[\hat{y^i} - y^i_{\lambda_{dbcs}}$]\;
  }
\end{algorithm}


Here the hyper-parameter $\lambda_{dbcs} > 0$ controls the trade-off between \textit{informativeness of the gradient} and \textit{faithfulness to the original function}.

\subsection{Answer and Explanation Selection}




The solved adjacency variable $\hat{y}^i$ represents the selected edges for each candidate answer choice $c_i$. Not all datasets provide gold explanations. Moreover, even when the gold explanations are available, they are only available for the correct answer with no explanations for the \textit{wrong} answer. 

In order to tackle these shortcomings and ensure end-to-end differentiability, we use the softmax ($\sigma$) of the objective score ($\hat{W}^i \cdot \hat{y}^i$) as the probability score for each candidate answer.

We multiply each objective score $\hat{W}^i \cdot \hat{y}^i$ value by the temperature hyperparameter ($T$) to obtain soft probability distributions $\gamma^i$ (where $\gamma^i = (\hat{W}^i \cdot \hat{y}^i)\cdot T$). The aim is for the correct answer $c_{ans}$ to have the highest probability.


In order to achieve this, we use the cross entropy loss $l_c$ as follows to calculate the answer selection loss $\mathcal{L}_{ans}$ as follows:
\begin{equation}
\begin{split}
    \mathcal{L}_{ans} = l_c(\sigma(\gamma^1,~\gamma^2,~\cdots~\gamma^n),~c_{ans})
\end{split}
\end{equation}





If gold explanations are available, we complement $\mathcal{L}_{ans}$ with explanation loss $\mathcal{L}_{exp}$. We employ binary cross entropy loss $l_b$ between the selected explanatory facts and gold explanatory facts $F_{exp}$ for the explanatory loss as follows:
\begin{equation}
    \mathcal{L}_{exp} = l_b(\hat{y}^{ans}[f_1,~f_2,~\dots,~f_k],~F_{exp})
\end{equation} 

We calculate the total loss ($\mathcal{L}$) as weighted by hyperparameters $\lambda_{ans}$, $\lambda_{exp}$ as follows:

\begin{equation}
    \mathcal{L} = \lambda_{ans}\mathcal{L}_{ans} + \lambda_{exp}\mathcal{L}_{exp}
\end{equation}




\section{Empirical Evaluation}

\subsection{Answer and Explanation Selection}

\begin{table*}[ht]
    \centering
    \resizebox{\textwidth}{!}{
    \begin{tabular}{p{4cm}ccccccP{2cm}}
    \toprule
    \multirow{3}{*}{\textbf{Model}}  & \multicolumn{6}{c}{\textbf{Explanation Selection (\textit{dev})}}  & \multirow{3}{2cm}{\centering\textbf{Answer Selection (\textit{test})}}\\
    \cmidrule{2-7}
     & \multicolumn{2}{c}{\textbf{Precision}} & \multicolumn{3}{P{3cm}}{\textbf{Explanatory Consistency}} & \multirow{2}{*}{\textbf{Faithfulness}} \\
         \cmidrule(l{2pt}r{2pt}){2-3} \cmidrule(l{2pt}r{2pt}){4-6}

     & \textbf{@2} & \textbf{@1} & \textbf{@3} & \textbf{@2} &  \textbf{@1}  \\
    \midrule
    \underline{\textbf{Baselines}} \\
    BERT$_{Base}$& - & - & - &- &- &-  & 45.43\\
    BERT$_{Large}$& - & - & - &- &- &- & 49.63\\
    \midrule
    
    Fact Retrieval (FR) Only& 30.19 & 38.49 & 21.42 &  15.69 & 11.64 & - & -\\
    BERT$_{Base}$ + FR&  - & - & - &- &- & 52.65 &58.06\\
    BERT$_{Large}$ + FR&    - & - & - &- &-  & 51.23 &59.32\\
    \midrule
    ExplanationLP &  40.41 & 51.99 & 29.04 & 14.14 & 11.79  & 71.11 &62.57\\
    \textit{Diff-}Explainer  & 41.91 & 56.77 & 39.04  & 20.64 & 17.01 & 72.22 &71.48\\
    \midrule
    \midrule
    \underline{\textbf{\textit{Diff-}Comb Explainer}}\\
    - Answer selection only & 45.75 & 61.01 &\underline{\textbf{49.04}} & 29.99 & 18.88 & 73.37 & 72.04\\
    - Answer and explanation selection  & \underline{\textbf{47.57}} & \underline{\textbf{63.23}} & 43.33 & \underline{\textbf{33.36}} & \underline{\textbf{20.71}} & \underline{\textbf{74.47}} &\underline{\textbf{73.46}}\\
    \bottomrule
    \end{tabular}}
    \caption{Comparison of explanation and answer selection of \textit{Diff-}Comb Explainer against other baselines. Explanation Selection was carried out on the \textit{dev} set as the \textit{test} explanation was not public available.}
    \label{tab:comb_worldtreee_answer_explanation}
\end{table*}

We use the WorldTree corpus~\citep{xie2020worldtree} for the evaluation of explanation and answer selection. The 4,400 questions and explanations in the WorldTree corpus are split into three different subsets: \emph{train-set}, \emph{dev-set} and \emph{test-set}. We use the \emph{dev-set} to assess the explainability performance since the explanations for \emph{test-set} are not publicly available. The background knowledge consists of 5000 abstract facts from the WorldTree table store (WTree)~\citep{xie2020worldtree} and over 100,000 \textit{is-a} grounding facts from ConceptNet~\citep{DBLP:journals/corr/SpeerCH16}.

\paragraph{Baselines:} We use the following baselines to compare against our approach for the WorldTree corpus:
\begin{enumerate}[noitemsep,leftmargin=*]
    \item \textbf{BERT$_{Base}$ and BERT$_{Large}$}~\citep{devlin2019bert}: To use BERT for this task, we concatenate every hypothesis with $k$ retrieved facts, using the separator token \texttt{[SEP]}. We use the HuggingFace~\citep{DBLP:journals/corr/abs-1910-03771} implementation of \textit{BertForSequenceClassification}, taking the prediction with the highest probability for the positive class as the correct answer. Notice that Diff-Comb Explainer adopts the same neural architecture for sentence representation and inference.
    
     \item \textbf{ExplanationLP}: Non-differentiable version of ExplanationLP. Using the constraints stated in Section~\ref{sec:comb_approach}, we fine-tune the $\theta$ parameters using Bayesian optimization and frozen STrans representations following \cite{thayaparan2021explainable}. This baseline aims to evaluate the impact of end-to-end fine-tuning over the non-differentiable solver.
     
     \item \textbf{\textit{Diff-}Explainer}: \textit{Diff-}Explainer has already exhibited better performance over existing ILP-based approaches including TableILP~\citep{khashabi2016question}, TupleILP~\citep{khot2017answering} and graph-based neural approach PathNet~\citep{kundu-etal-2019-exploiting} and represents the current state-of-the-art combinatorial solver in the field. To allows for a direct comparison with our approach, we use the ExplanationLP constraints with \textit{Diff-}Explainer. We adopt the same set of hyperparameters and knowledge base reported in \citet{diff-explainer}.
    
\end{enumerate}



\paragraph{Evaluation Metrics.} The answer selection is evaluated using accuracy. For evaluation of explanation selection, we use Precision$@K$. In addition to Precision$@K$, we introduce two new metrics to evaluate the truthfulness of the answer selection to the underlying inference. The metrics are as follows:



\paragraph{Faithfulness} The aim of faithfulness is to measure the extent to which correct answer predictions are indeed derived from correct explanations and, vice-versa, wrong answer predictions are derived from wrong explanations. Given the set of correct answers $A_{Q_c}$, wrong answers $A_{Q_w}$, answers with at least one correctly retrieved explanation sentence $A_{Q_1}$ and set of questions where no correct explanation fact is retrieved $A_{Q_0}$, the faithfulness score is defined as follows:
    \begin{equation}
      \frac{\vert A_{Q_w} \cap A_{Q_0} \vert +  \vert A_{Q_c} \cap A_{Q_1} \vert}{\vert A_{Q_c}  \cup  A_{Q_w} \vert}
    \end{equation}

A higher faithfulness score implies, therefore, that the underlying inference performed by the model is more in line with the final answer prediction.

\paragraph{Explanation Consistency$@K$} As similar questions require similar underlying inference~\citep{valentino2021unification,valentino2022hybrid}, we expect expect the explanations to reflect the similarity between test instances. Specifically, given a test question $Q_t$ and the explanation constructed by the model $E_t$, we select the set of Questions $Q^s_t= \{ Q^1_t,~Q^2_t,~\dots\}$ that have at least $K$ gold explanatory facts overlapping with the gold explanation for $Q_t$, $E^s_t= \{ e^1_t,~e^2_t,~\dots\}$. Given this premise, we measure the explanation consistency$@K$ as follows:
    \begin{equation}
        \frac{\sum_{e^i_t \in E^s_t}[e^i_t \in E_t]}{\sum_{e^i_t \in E^s_t} \vert e^i_t \vert}
    \end{equation}
Specifically, explanation consistency measures the percentage of overlapping explanatory facts retrieved by the models between hypotheses that share at least $K$ similar gold explanations. Higher explanatory consistency, therefore, implies that the reasoning across similar questions tends to produce similar explanations.

\subsection{Results}
Table~\ref{tab:comb_worldtreee_answer_explanation} illustrates the explanation and answer selection performance of \textit{Diff-}Comb Explainer and the baselines.  We report scores for \textit{Diff-}Comb Explainer trained for only the answer and optimised jointly for answer and explanation selection.
Since BERT does not provide explanations, we use facts retrieved from the fact retrieval for the best $k$ configuration ($k=3$) as explanations. We also report the scores for BERT without explanations.

We draw the following conclusions from the results obtained in Table~\ref{tab:comb_worldtreee_answer_explanation} (The performance increase here are  expressed in absolute terms):

\noindent \textbf{(1)} \textit{Diff-}Comb Explainer improves answer selection performance over the non-differentiable solver by 9.47\% when optimised only on answer selection and  10.89\% when optimised on answer and explanation selection. This observation highlights the impact of the end-to-end differentiable framework. We can also observe that strong supervision on explanation selection yields better performance than weak supervision on answer selection only.

\noindent \textbf{(2)} \textit{Diff-}Comb Explainer outperforms the best BERT model by 14.14\% on answer selection. This increase in performance demonstrates that integrating constraints with Transformer-based encoders can lead to better NLI performance. 

\noindent \textbf{(3)} \textit{Diff-}Comb Explainer outperform the best \textit{Diff-}Explainer configuration (answer and explanation selection) by 0.56\% even in the weak supervision setting (answer only optimization). We also outperform  \textit{Diff-}Explainer by 1.98\% in the best setting. 
    
\noindent \textbf{(4)}  \textit{Diff-}Comb Explainer shows superior performance in selecting relevant explanations over existing combinatorial solvers. \textit{Diff-}Comb Explainer outperforms the non-differentiable solver at Precision@K by 8.41\% ($k=$1) and 6.05\% ($k=$2). Moreover, the proposed approach outperforms \textit{Diff-}Explainer by 3.63\% ($k=$1) and 4.55\% ($k=$2). This demonstrates the impact of preserving the original ILP formulation on downstream inference. Finally, the improvement of Precision$@K$ over the Fact Retrieval only (demonstrated with BERT + FR) by 16.98\% ($k=$1) and 24.74\% ($k=$2) underlines the robustness of our approach to noise propagated by the upstream fact retrieval.
    
\noindent \textbf{(5)} Diff-Comb Explainer exhibits higher Explanation Consistency over the other solvers. This performance shows that the optimization model is learning and applying consistent inference across similar instances. We also outperform the fact retrieval model based on Transformers and fine-tuned on gold explanations.
    
\noindent \textbf{(6)} Answer prediction by \textit{black-box} models like BERT do not reflect the explanation provided. This fact is indicated by the low Faithfulness score obtained by both BERT$_{Base}$/BERT$_{Large}$. In contrast, the high constraint-based solver's Faithfulness scores emphasise how the underlying inference reflects on the final prediction. In particular, our approach performs better than the non-differentiable models and \textit{Diff-}Explainer.

In summary, despite the fact that \textit{Diff-}Explainer and \textit{Diff-}Comb Explainer adopts the same underlying set of constraints, the results demonstrate that our model yields better performance, indicating that accurate predictions generated by ILP solvers reflect in better inference performance than sub-optimal counterparts. 

\subsection{Comparing Answer Selection with ARC Baselines}
\label{sec:answer_selection_blackbox}

\begin{table}[t]
		 \resizebox{\linewidth}{!}{
     \begin{tabular}{@{}>{\raggedright}p{4.6cm}cc@{}}
    \toprule
    \textbf{Model} & \textbf{Explainable}   & \textbf{Accuracy}  \\
    \midrule
       BERT$_{Large}$ & No  & 35.11 \\ 
     \midrule
     IR Solver~\citep{clark2016combining} & Yes & 20.26\\
     TupleILP~\citep{khot2017answering} & Yes  & 23.83   \\ 
     TableILP~\citep{khashabi2016question} & Yes   &  26.97  \\ 
    ExplanationLP\newline~\citep{thayaparan2021explainable} & Yes & 40.21 \\
     DGEM~\citep{clark2016combining} & Partial  &27.11  \\ 
     KG$^{2}$~\citep{zhang2018kg} & Partial   &  31.70 \\
     ET-RR~\citep{ni2018learning} & Partial & 36.61 \\
     Unsupervised AHE~\citep{yadav2019alignment} & Partial    &33.87 \\ 
     Supervised AHE~\citep{yadav2019alignment} & Partial  & 34.47 \\ 
     AutoRocc~\citep{yadav-etal-2019-quick} & Partial   & 41.24 \\ 
    \textit{Diff-}Explainer (ExplanationLP)~\citep{diff-explainer} & Yes  &\textbf{42.95}  \\ 
    \midrule
    \textit{Diff-}Comb Explainer (ExplanationLP) & Yes  &\textbf{43.21}  \\ 
     \bottomrule
    \end{tabular}
    }
    \caption{\small ARC challenge scores compared with other Fully or Partially explainable approaches trained \textit{only} on the ARC dataset.}
    \label{tab:comb_arc_baselines}
\end{table}

\begin{table*}[ht!]
    \centering
    \small
    \resizebox{\textwidth}{!}{\begin{tabular}{p{14cm}}
    \toprule
    \textbf{Question (1):} Which measurement is best expressed in light-years?: 
    \textbf{Correct Answer:} the distance between stars in the Milky Way. \\
    \underline{ExplanationLP} \\
    \textbf{Answer}: the time it takes for planets to complete their orbits.~\textbf{Explanations}: \textit{(i)} a complete revolution; orbit of a  planet around its star takes 1; one planetary year, \textit{(ii)} a light-year is used for describing long distances \\
    \underline{\textit{Diff-}Explainer} \\
    \textbf{Answer}: the time it takes for planets to complete their orbits.~\textbf{Explanations}: \textit{(i)} a light-year is used for describing long distances, \textit{(ii)} light year is a measure of the distance light travels in one year \\
    \underline{\textit{Diff-}Comb Explainer}  
    \\
    \textbf{Answer}: the distance between stars in the Milky Way.~\textbf{Explanations}: \textit{(i)} light years are a  astronomy unit used for measuring length, \textit{(ii)} stars are located light years apart from each other \\
    \midrule
    
    \textbf{Question (2):} Which type of precipitation consists of frozen rain drops?: 
    \textbf{Correct Answer:} sleet. \\
    \underline{ExplanationLP} \\
    \textbf{Answer}: snow.~\textbf{Explanations}: \textit{(i)} precipitation is when snow fall from clouds to the Earth, \textit{(ii)} snow falls \\
    \underline{\textit{Diff-}Explainer} \\
    \textbf{Answer}: sleet.~\textbf{Explanations}: \textit{(i)} snow falls, \textit{(ii)} precipitation is when water falls from the sky \\
    \underline{\textit{Diff-}Comb Explainer}  
    \\
    \textbf{Answer}: sleet.~\textbf{Explanations}: \textit{(i)} sleet is when raindrops freeze as they fall, \textit{(ii)} sleet is made of ice \\
    \midrule

    \textbf{Question (3):} Most of the mass of the atom consists of?: 
    \textbf{Correct Answer:} protons and neutrons. \\
    \underline{ExplanationLP} \\
    \textbf{Answer}: neutrons and electrons.~\textbf{Explanations}: \textit{(i)} neutrons have more mass than an electron, \textit{(ii)} neutrons have more mass than an electron \\
    \underline{\textit{Diff-}Explainer} \\
    \textbf{Answer}: protons and neutrons.~\textbf{Explanations}: \textit{(i)} the atomic mass is made of the number of protons and neutrons, \textit{(ii)} precipitation is when water falls from the sky \\
    \underline{\textit{Diff-}Comb Explainer}  \\
   \textbf{Answer}: protons and neutrons.~\textbf{Explanations}: \textit{(i)} the atomic mass is made of the number of protons and neutrons, \textit{(ii)} precipitation is when water falls from the sky \\

    \bottomrule
    \end{tabular}}
    \caption{Example of predicted answers and explanations obtained from our model with different combinatorial solvers.}
    \label{tab:comb_qa_examples_results}
\end{table*}


Table~\ref{tab:comb_arc_baselines} presents a comparison of publicly reported baselines on the ARC Challenge-Corpus~\citep{clark2018think} and our approach. These questions have proven to be challenging to answer for other LP-based NLI and neural approaches. 

In Table~\ref{tab:comb_arc_baselines}, to provide a fair comparison, we list the models that have been trained \textit{only} on the ARC corpus. Here, the explainability column indicates if the model delivers an explanation for the predicted answer. A subset of the models produces evidence for the answer but remains intrinsically black-boxes. These models have been marked as \textit{Partial}. As illustrated in the Table~\ref{tab:comb_arc_baselines}, \textit{Diff-}Comb Explainer outperforms the best non-differentiable ILP solver model (ExplanationLP) by 2.8\%. We also outperform a transformer-based model (AutoRocc) by 1.97\%. While our improvement over \textit{Diff-}Explainer is small, we still demonstrate performance improvements for answer selection. 

\subsection{Qualitative Analysis}


Table~\ref{tab:comb_qa_examples_results} illustrates some of the explanations constructed by ExplanationLP, \textit{Diff-}Explainer and \textit{Diff-}Comb Explainer. Both explanations and answer predictions in Question (1) are entirely correct for our model. In this example, both ExplanationLP and \textit{Diff-}Explainer have failed to retrieve any correct explanations or predict the correct answer. Both the approaches are distracted by the strong lexical overlaps with the wrong answer.

In Question (2) Diff-Comb Explainer produces the correct answer with at least one explanation sentence marked as correct. Here,  \textit{Diff-}Explainer provides the correct answer prediction with both the retrieved facts marked as incorrect showing a lower faithfulness compared to our proposed approach.

In Question (3) both our model and \textit{Diff-}Explainer provide the correct answer supported by an incorrect explanation. While, the aforementioned qualitative (Question 1 and 2) and quantitative measures (Explanatory Consistency$@K$, Faithfulness) indicate the superiority of Diff-Comb Explainer in generating supporting explanations, we still find cases where incorrect inference can lead to the correct answer. In most of the cases, we found that the inference is distracted by strong lexical overlaps with the question.

In general, however, we can conclude that Diff-Comb Explainer is more robust when compared to \textit{Diff-}Explainer and existing ILP models.

\begin{figure}[ht!]
    \centering    \includegraphics[width=\columnwidth]{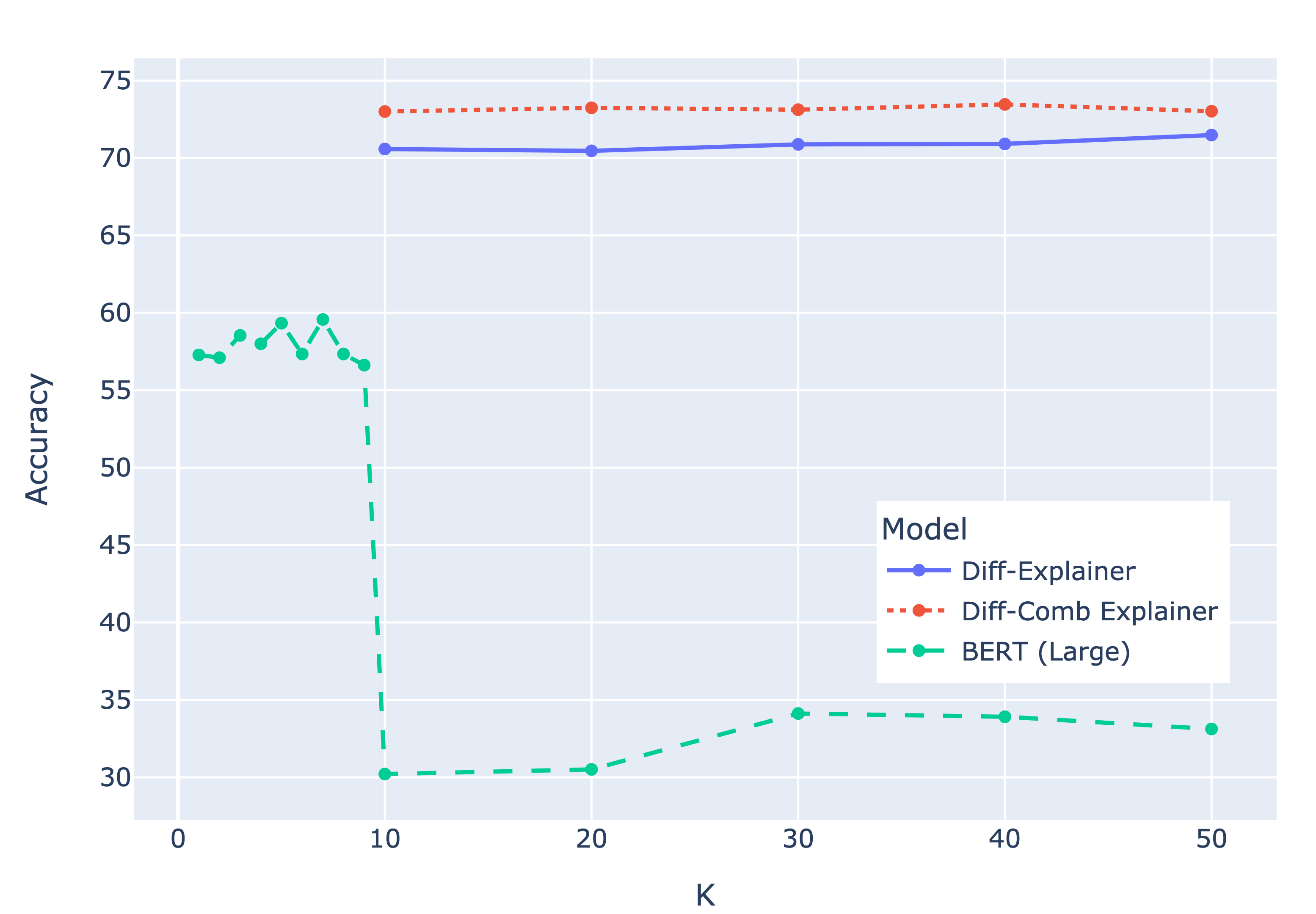}
    \caption{Comparison of accuracy for different number of retrieved facts.}
    \label{fig:comb_k_facts_plot}
\end{figure}

\subsection{Knowledge aggregation with increasing distractors}

One of the key characteristics identified by~\citet{diff-explainer} is the robustness of \textit{Diff-}Explainer to distracting noise. In order to evaluate if our model also exhibits the same characteristics, we tested our model with an increasing number of retrieved facts $k$ from the KB and plot the resulting answer accuracy in Figure~\ref{fig:comb_k_facts_plot}. 

As illustrated in the Figure, similar to \textit{Diff-}Explainer, our approach performance remains stable with increasing distractors. We also continue to outperform \textit{Diff-}Explainer across all sets of $k$.

BERT performance, on the other side, drops drastically with increasing distractors. This phenomenon is in line with existing findings~\citep{diff-explainer,yadav-etal-2019-quick}. We hypothesise that with increasing distractors, BERT overfits quickly with spurious inference correlation. On the other hand, our approach circumvents this problem with the inductive bias provided by the constraint optimization layer.

\section{Conclusion}

This paper proposed a novel framework for encoding explicit and controllable assumptions as part of an end-to-end learning framework for explanation-based NLI using Differentiable Blackbox Combinatorial Solvers~\citep{poganvcic2019differentiation}. We empirically demonstrated improved answer and explanation selection performance compared with the existing differentiable constraint-based solvers~\citep{diff-explainer}. Moreover, we demonstrated performance gain and increased robustness to noise when combining precise constraints with transformer-based architectures. In this paper, we adopted a specific set of constraints to ensure fairness in the evaluation, but, in principle, it is possible to extend the proposed architecture with different inference constraints.

\textit{Diff-}Comb Explainer improves upon recent work by ~\citet{diff-explainer} and investigates the combination of symbolic knowledge (expressed via constraints) with neural representations. We hope this work will encourage researchers to encode different domain-specific priors, leading to more robust, transparent and controllable neuro-symbolic inference models for NLP.

\section{Limitations}

The ILP formulation is NP-complete. Therefore, as the number of variables in the problem increases the computational complexity also increases exponentially. However, this challenge can be alleviated with strong solvers and the relaxation of some constraints. The performance of the proposed approach is not yet comparable with the scores achieved by Large Language Models (LLMs) ~\cite{khashabi2020unifiedqa}. However, it should be noted that, unlike LLMs, the proposed approach was trained only on the dataset it was tested on and exhibited higher performance compared to the same Transformer encoders integrated in the architecture. These types of neuro-symbolic models can potentially lead to smaller, domain-specific models that require little training data for more efficient and transparent natural language inference.

\section*{Acknowledgements}
This work was partially funded by the Swiss National Science Foundation (SNSF) project NeuMath (\href{https://data.snf.ch/grants/grant/204617}{200021\_204617}), by the EPSRC grant EP/T026995/1 entitled “EnnCore: End-to-End Conceptual Guarding of Neural Architectures” under Security for all in an AI enabled society, by the CRUK National Biomarker Centre, and supported by the Manchester Experimental Cancer Medicine Centre.
The authors would like to thank the anonymous reviewers and editors of LREC-COLING for their constructive feedback. Additionally, we would like to thank the Shared Facilities of the University of Manchester and Idiap Research Institute for providing the infrastructure to run our experiments.

\section{Bibliographical References}
\bibliographystyle{lrec-coling2024-natbib}
\bibliography{refs2}


\end{document}